\documentclass{article}

\PassOptionsToPackage{hyperfootnotes=false}{hyperref}
\usepackage[preprint]{corl_2026} 

\title{FlashVLA: Streaming Action Decoding for \\ Fast and Asynchronous VLA Inference}

\usepackage{color,xcolor}
\usepackage{epsfig}
\usepackage{graphicx}

\usepackage{adjustbox}
\usepackage{array}
\usepackage{booktabs}
\usepackage{colortbl}
\usepackage{float,wrapfig}
\usepackage{hhline}
\usepackage{multirow}
\usepackage{subcaption} 
\usepackage[toc,page]{appendix}
\usepackage{makecell}
\usepackage{tabularx}

\usepackage{amsmath,amsfonts,amsthm,amssymb}
\usepackage{bm}
\usepackage{nicefrac}
\usepackage{microtype}
\usepackage{inconsolata}
\usepackage{pifont}
\usepackage{relsize}

\usepackage{changepage}
\usepackage{extramarks}
\usepackage{fancyhdr}
\usepackage{lastpage}
\usepackage{setspace}
\usepackage{soul}
\usepackage{xspace}
\usepackage{indentfirst}
\usepackage{paralist}
\usepackage{booktabs,arydshln}

\usepackage{algorithm, algpseudocode}
\usepackage{enumerate}
\usepackage{lipsum}
\usepackage{minted}
\usepackage{pgfplots}
\pgfplotsset{compat=1.18}

\newcolumntype{L}[1]{>{\raggedright\let\newline\\\arraybackslash\hspace{0pt}}m{#1}}
\newcolumntype{C}[1]{>{\centering\let\newline\\\arraybackslash\hspace{0pt}}m{#1}}
\newcolumntype{R}[1]{>{\raggedleft\let\newline\\\arraybackslash\hspace{0pt}}m{#1}}

\newcommand{\tablestyle}[2]{\setlength{\tabcolsep}{#1}\renewcommand{\arraystretch}{#2}\centering\footnotesize}

\newcommand{\ignorethis}[1]{}

\renewcommand*{\thefootnote}{\fnsymbol{footnote}}

\makeatletter
\DeclareRobustCommand\onedot{\futurelet\@let@token\@onedot}
\def\@onedot{\ifx\@let@token.\else.\null\fi\xspace}

\makeatother

\makeatletter
\def\adl@drawiv#1#2#3{%
        \hskip.5\tabcolsep
        \xleaders#3{#2.5\@tempdimb #1{1}#2.5\@tempdimb}%
                #2\z@ plus1fil minus1fil\relax
        \hskip.5\tabcolsep}
\newcommand{\cdashlinelr}[1]{%
  \noalign{\vskip\aboverulesep
           \global\let\@dashdrawstore\adl@draw
           \global\let\adl@draw\adl@drawiv}
  \cdashline{#1}
  \noalign{\global\let\adl@draw\@dashdrawstore
           \vskip\belowrulesep}}
\makeatother

\definecolor{citecolor}{HTML}{0071bc}
\definecolor{mydarkblue}{rgb}{0,0.08,1}
\definecolor{mydarkgreen}{rgb}{0.02,0.6,0.02}
\definecolor{mydarkred}{rgb}{0.8,0.02,0.02}
\definecolor{mydarkorange}{rgb}{0.40,0.2,0.02}
\definecolor{mypurple}{RGB}{111,0,255}
\definecolor{myred}{rgb}{1.0,0.0,0.0}
\definecolor{mygold}{rgb}{0.75,0.6,0.12}
\definecolor{mydarkgray}{rgb}{0.66, 0.66, 0.66}

\definecolor{darkblue}{rgb}{0,0.08,1}
\definecolor{darkgreen}{rgb}{0.02,0.6,0.02}
\definecolor{darkred}{rgb}{0.8,0.02,0.02}
\definecolor{darkorange}{rgb}{0.40,0.2,0.02}
\definecolor{darkpurple}{RGB}{111,0,255}

\definecolor{mydarkblue}{rgb}{0,0.08,1}

\def\method{FlashVLA\xspace}

\definecolor{mygray}{gray}{0.93}
\definecolor{mygreen}{HTML}{009000}
\definecolor{myred}{HTML}{ea4335}

\author{
\textbf{Zekai Li}\textsuperscript{1} \quad
\textbf{Jiaming Tang}\textsuperscript{2} \quad
\textbf{Zhijian Liu}\textsuperscript{1}\\
$^1$UC San Diego \quad $^2$MIT \\
\\ \url{https://github.com/z-lab/flashvla.git}
}

\begin{document}
\maketitle
\renewcommand{\thefootnote}{\arabic{footnote}}


\begin{abstract}

Vision-Language-Action (VLA) models are increasingly promising for robotic manipulation, yet their real-world deployment remains bottlenecked by high inference latency and unstable asynchronous execution. This challenge is particularly pronounced in flow-matching-based VLA models, where action decoding requires multiple iterative steps conditioned on the VLM context. While efficient inference methods improve control frequency and asynchronous methods reduce execution idle time, existing approaches often fail to jointly achieve low-latency inference and accurate, temporally consistent asynchronous execution. We introduce \textbf{FlashVLA}, a streaming action decoding framework that addresses both challenges in a unified formulation. FlashVLA maintains a streaming action buffer with multiple chunks at different noise levels and decodes them using chunk-wise causal attention. This design allows FlashVLA to produce one executable action chunk per inference step. Moreover, its chunk-wise autoregressive formulation implicitly preserves action continuity, enabling smooth asynchronous execution without extra future-state conditioning. Across extensive simulated and real-world experiments, FlashVLA substantially improves inference speed while maintaining strong task performance. It can achieve $\geq$30\,Hz control frequency on a single GPU with smooth asynchronous inference in real-world deployment.

\end{abstract}


\section{Introduction}
\label{sec:intro}

\begin{figure}[t]
\centering
\begin{subfigure}[t]{0.75\linewidth}
    \centering
    \includegraphics[width=\linewidth]{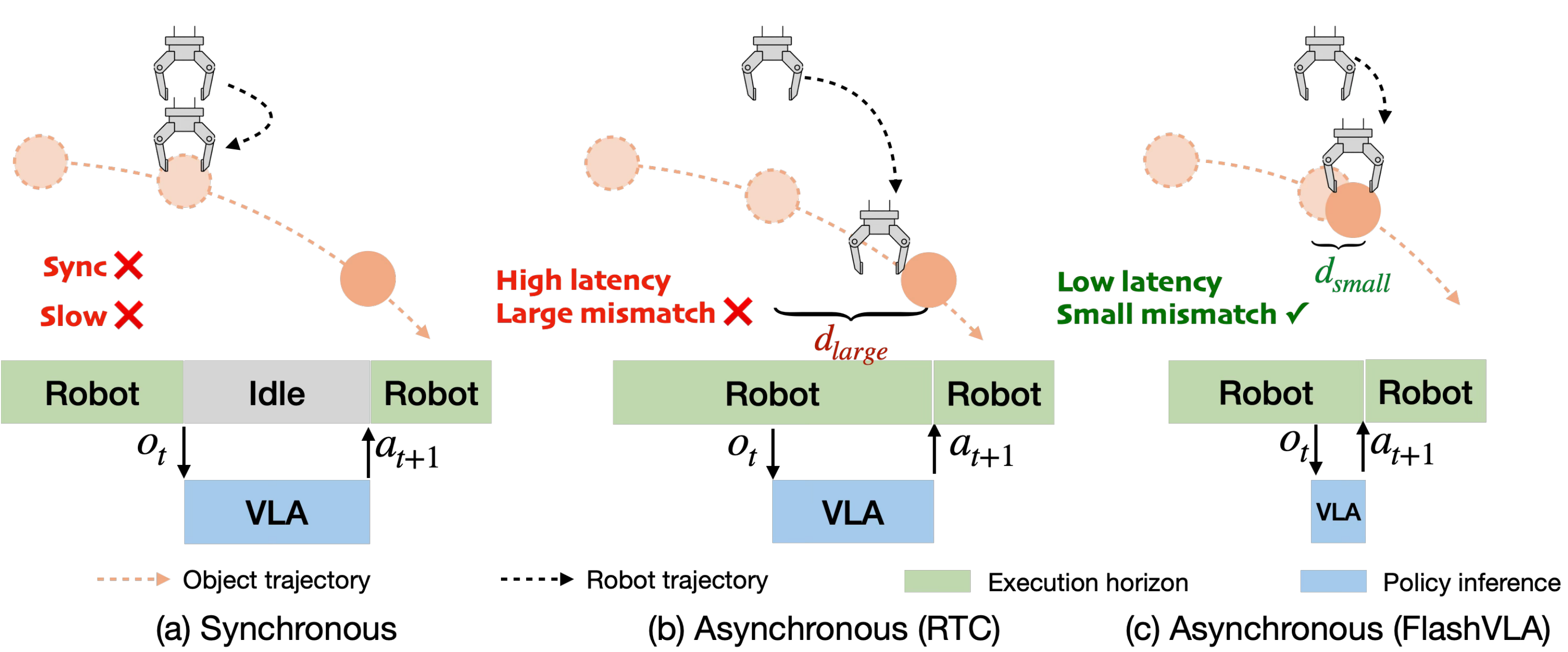}
    \caption{Conceptual illustration of prior VLA inference: it either stalls the robot at every chunk boundary (synchronous) or predicts from a stale observation whose mismatch grows with the lookahead (asynchronous with future-state conditioning). \method removes both costs at once by decoding chunks jointly under chunk-wise causal attention.}
    \label{fig:overview}
\end{subfigure}
\hfill
\begin{subfigure}[t]{0.23\linewidth}
    \centering
    \includegraphics[width=\linewidth]{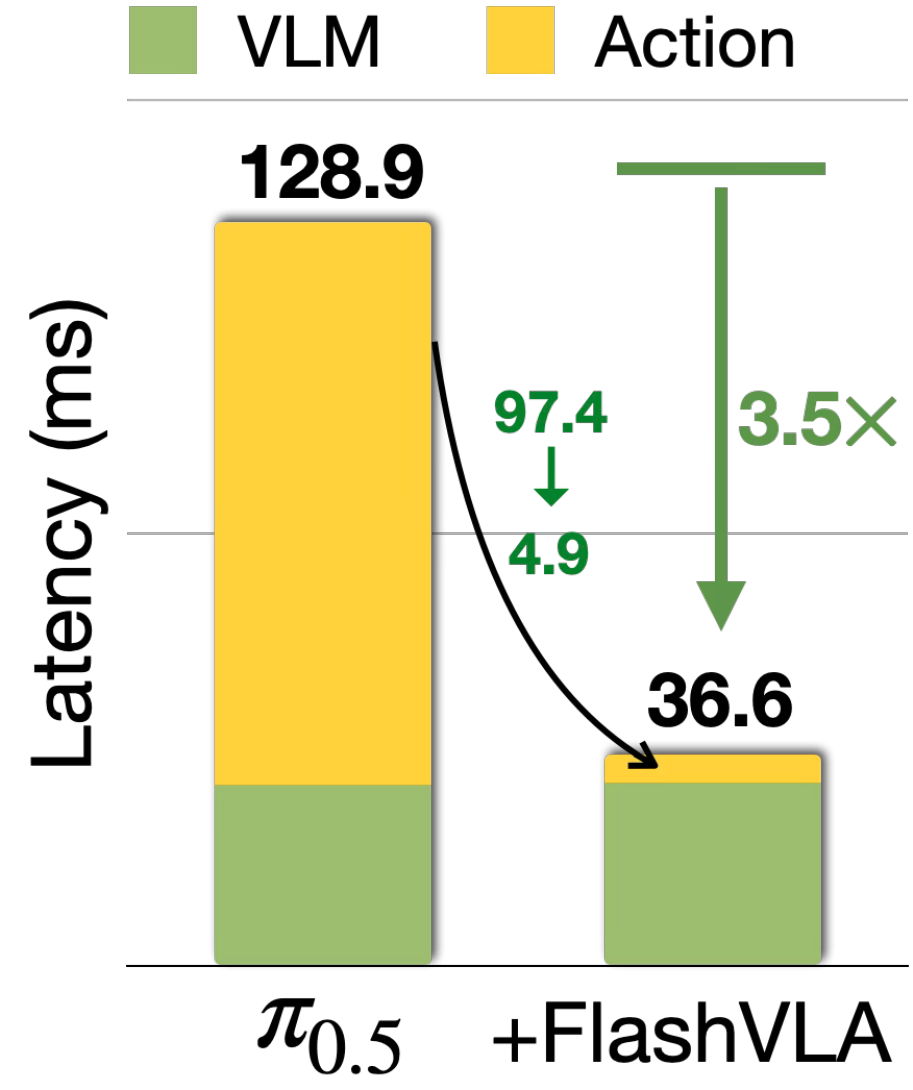}
    \caption{Action decoding consumes 75\% of inference time. \method targets this bottleneck, reaching 3.5$\times$ overall speedup.}
    \label{fig:profile}
\end{subfigure}
\caption{(a) Two prevailing VLA inference patterns and \method's joint solution. (b) Per-policy-invocation profile of $\pi_{0.5}$ and \method. Action decoding is the dominant bottleneck.}
\end{figure}

Vision-Language-Action (VLA) models~\citep{openvla, gr00t, pi0, lingbot-vla} map visual observations and language instructions directly to robot actions, and have become a leading paradigm for end-to-end manipulation. Recent models such as $\pi_{0.5}$~\citep{pi05} couple pretrained vision-language backbones with a flow-matching action expert, achieving strong generalization across diverse tasks. Yet their inference latency obstructs real-time control: profiling $\pi_{0.5}$ shows that action decoding alone consumes \textbf{75\%} of per-step inference time (Fig.~\ref{fig:profile}), driven by the ten sequential denoising steps required per chunk. The profile is measured on an RTX 4090 with two views, with no system-level optimization applied to either model.

This bottleneck creates a dilemma for VLA deployment. Synchronous inference keeps observations and actions temporally aligned, but stalls the robot at every chunk boundary while the next chunk is decoded. Asynchronous inference~\citep{vlash, train-rtc, streamingvla} overlaps prediction with execution, but predicts from a stale observation; the actions arrive at a robot state the model never saw, and the longer the lookahead, the larger the temporal mismatch. Prior work addresses one side at a time. Efficient-inference methods reduce per-step latency through lightweight architectures~\citep{smolvla}, compression~\citep{qvla}, or token pruning~\citep{efficientvla}, while asynchronous methods condition on future states or planned actions~\citep{vlash, train-rtc} to suppress mismatch. Neither closes the gap: a faster synchronous schedule still stalls at chunk transitions, and a smoother asynchronous schedule still requires the lookahead that re-amplifies mismatch.

\textbf{Our central observation is that these two failure modes share a single cause.} Flow-matching VLAs decode each action chunk in \emph{isolation}: from pure noise, conditioned only on the current observation, and decoupled from any chunks already in flight. Isolation concentrates all ten denoising steps inside a single chunk decode (the latency cost) and leaves each new chunk uninformed about the trajectory it is about to join (the mismatch cost). Both costs collapse under a single structural change: decode chunks \emph{jointly} rather than in isolation.

We introduce \textbf{\method}, a streaming action decoding framework that realizes this idea. Inspired by streaming chunk-wise diffusion for long video~\citep{diffusio_forcing, streamdit, magi1}, \method maintains a streaming buffer of action chunks at \emph{staggered} noise levels and advances them jointly under chunk-wise causal attention. Two benefits follow from this single mechanism. First, each forward pass advances every chunk in the buffer by one denoising step, so after warm-up one executable chunk is emitted per step; each chunk passes through all $N$ denoising steps, amortized across time rather than concentrated in a single decode. Per-step action-decoding latency drops by up to \textbf{20$\times$}.\footnote{The ten-to-one pass reduction yields a 9.3$\times$ reduction in GPU work; eliminating inter-pass kernel-launch serialization raises the measured wall-clock speedup to 19.9$\times$.} Second, higher-noise future chunks attend to lower-noise near-execution chunks under the causal mask, so the model decoding for a future robot state is implicitly conditioned on the trajectory that state will arrive on; asynchronous continuity is recovered without any explicit future-state predictor or action-conditioning module.

Across LIBERO, RoboTwin~2.0, and real-world Franka tasks, \method matches or improves $\pi_{0.5}$ task success under one-step asynchronous delay, achieves up to a \textbf{2.43$\times$} end-to-end per-action speedup, and sustains \textbf{$\geq$30\,Hz} smooth control on a single GPU.

\section{Related Work}
\label{sec:related}

\paragraph{Efficient VLA inference.}
Most efforts to accelerate VLA inference make each forward pass cheaper but leave the iterative decoding loop intact: lightweight backbones~\citep{tinyvla, smolvla, evo1}, weight and activation compression~\citep{bitvla, efficientvla, qvla}, visual-token pruning~\citep{specprune-vla, vla-cache, lightvla}, and faster action heads~\citep{fast, streamingvla, faster}. A separate direction shortens the loop itself via few-step distillation of diffusion or flow-matching policies~\citep{consistency, shortcut}, at the cost of retraining and a tradeoff between sample quality and step count. \method changes \emph{what} the loop produces per pass: it is orthogonal to the first family and offers an alternative to the second that avoids step-count distillation.

\paragraph{Asynchronous VLA execution.}
Asynchronous inference~\citep{rtc, train-rtc, vlash, streamingvla, realtime-vla} hides decoding latency by overlapping prediction with execution, but predicts from a stale observation while the robot is at a state the model never saw. Prior work patches this mismatch \emph{after} decoding: by conditioning the next chunk on a predicted future state~\citep{vlash}, on previously planned actions~\citep{rtc}, or via training-time augmentation~\citep{train-rtc}. The patch cost grows with lookahead and requires auxiliary supervision or architectural surgery. Underlying all of these is an unstated assumption: action chunks are decoded in \emph{isolation}, leaving each one uninformed about the trajectory it is about to join. \method removes the assumption rather than treating its symptom. With chunk-wise causal attention, future chunks attend to the near-execution chunks already in flight, making asynchronous continuity a structural property of the decoder rather than an external correction.

\paragraph{Streaming chunk-wise diffusion.}
Long-video generation has converged on streaming chunk-wise diffusion: a buffer of clips at staggered noise levels is denoised jointly with causal masking from clean to noisy, emitting one clean clip per pass~\citep{diffusio_forcing, streamdit, magi1}. \method brings this paradigm to robot action decoding, where it is in fact a closer fit than video. Lower-noise chunks in the buffer are not just about to be \emph{shown} but about to be \emph{executed}, so attending to them implicitly conditions higher-noise future chunks on the trajectory the robot is actually following, and recovers asynchronous continuity without explicit future-state prediction. The adaptation to action decoding lies in matching buffer depth and noise schedule to the control-frequency target, and in applying causal masking at chunk rather than token granularity. To our knowledge \method is the first VLA to adopt this formulation. Across all three directions above, the same structural choice (decoding chunks jointly rather than in isolation) collapses the latency cost and the asynchronous-mismatch cost into a single fix.

\section{Method}
\label{sec:method}

\begin{figure}[t]
\centering

\begin{subfigure}[t]{0.73\linewidth}
    \centering
    \includegraphics[width=\linewidth]{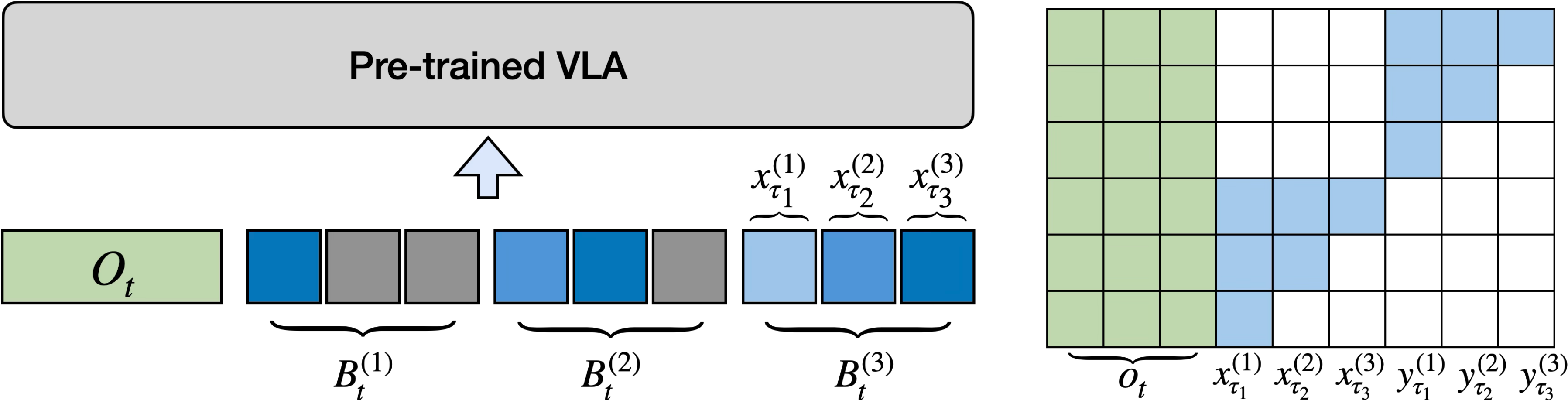}
\end{subfigure}
\hfill
\begin{subfigure}[t]{0.23\linewidth}
    \centering
    \includegraphics[width=\linewidth]{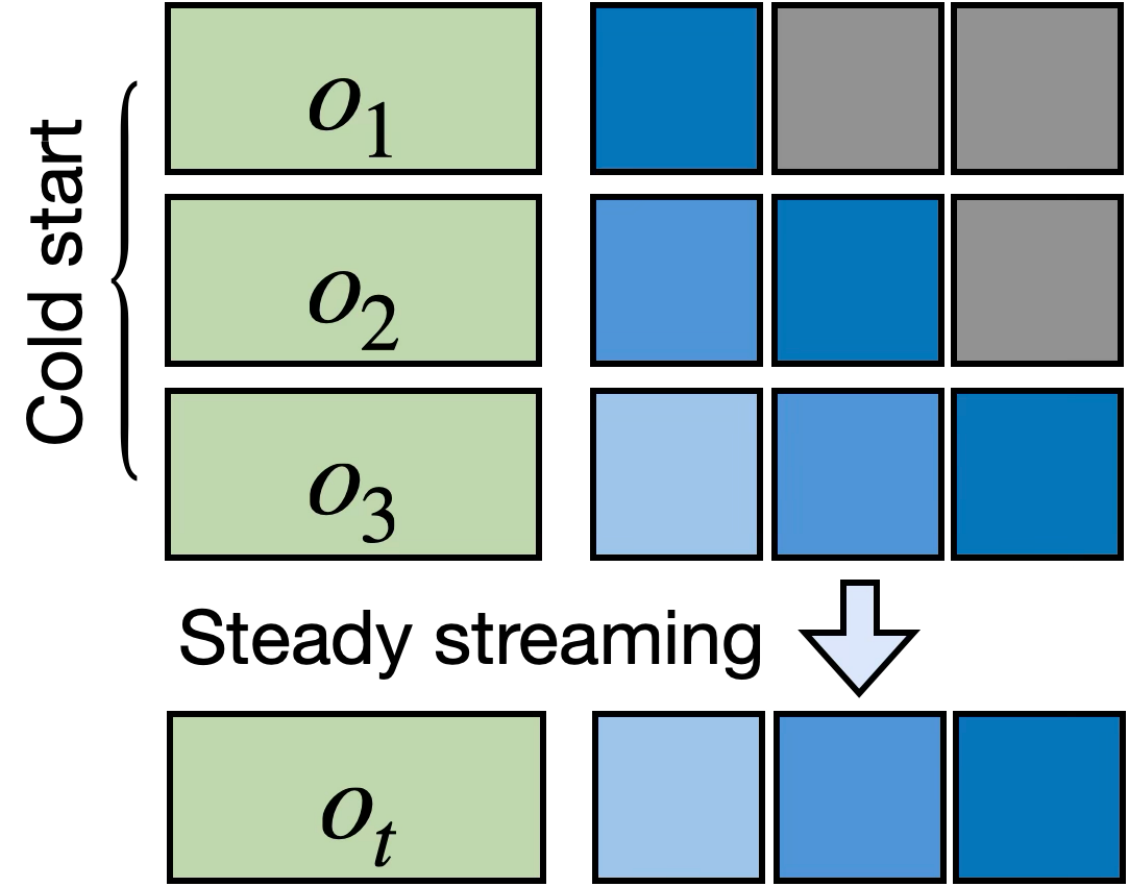}
\end{subfigure}
\caption{Overview of \method. \textbf{Left}: Multi-buffer joint fine-tuning packs all cold-start configurations for one observation into a single sample. Attention masks isolate buffers, and chunk-wise causal attention lets noisier chunks attend only to cleaner ones. \textbf{Right}: Streaming inference runs the buffer as a queue: cold-start fills it with $N{-}1$ warm-up passes, while steady streaming advances all chunks, executes the cleanest one, and appends fresh noise. Gray squares denote padding.}
\label{fig:method}
\end{figure}

We introduce \method, a streaming action decoding framework that realizes the central idea from \S\ref{sec:intro}: decode action chunks \emph{jointly} rather than in isolation. \method is a drop-in modification to any flow-matching-based VLA model, requiring only light architectural changes and a fine-tuning pass. We describe the chunk-wise autoregressive formulation (\S\ref{sec:formulation}), the streaming inference algorithm (\S\ref{sec:inference}), and the multi-buffer joint fine-tuning procedure used to adapt pretrained VLA models (\S\ref{sec:training}). Figure~\ref{fig:method} gives an overview of both the training and inference pipelines.

\subsection{Chunk-wise Autoregressive Formulation}
\label{sec:formulation}

A flow-matching VLA produces a continuous action chunk $\mathbf{a}_t = \{a_t, a_{t+1}, \ldots, a_{t+H-1}\}$ from observation $o_t$ by iteratively denoising a Gaussian sample $\mathbf{x}_1 \sim \mathcal{N}(0, I)$. The model learns a velocity field $v_\theta$ that transports noisy actions toward clean ones:
\begin{equation}
\mathbf{x}_\tau = (1 - \tau)\mathbf{a}_t + \tau \mathbf{z}, \quad \mathbf{z} \sim \mathcal{N}(0, I), \quad \tau \in [0, 1], \quad \frac{d\mathbf{x}_\tau}{d\tau} = \mathbf{z} - \mathbf{a}_t = v_\theta(\mathbf{x}_\tau, \tau \mid o_t).
\end{equation}
At inference, each chunk decode requires multiple sequential forward passes through the action expert (ten in $\pi_{0.5}$), all conditioned on the same observation $o_t$, and no chunk attends to any other. These two properties are the isolation identified in \S\ref{sec:intro}: concentrating all denoising inside a single chunk decode incurs the latency cost, and the lack of inter-chunk attention incurs the mismatch cost.

Rather than dedicating all denoising passes to a single chunk, \method maintains a buffer of $N$ chunks held at \emph{staggered} noise levels:
\begin{equation}
\begin{aligned}
\mathbf{B}_t &= [\mathbf{x}^{(1)}_{\tau_1}, \mathbf{x}^{(2)}_{\tau_2}, \ldots, \mathbf{x}^{(N)}_{\tau_N}], \quad \tau_1 < \tau_2 < \cdots < \tau_N, \\
u_i &\sim \operatorname{Beta}(1.5, 1.0), \quad \tau_i = 0.001 + 0.999\,\frac{i-1+u_i}{N}.
\end{aligned}
\end{equation}
The first slot holds an almost-clean chunk about to execute; the last slot holds a pure-noise chunk that will execute several steps later. A single forward pass advances every chunk by one step, and after warm-up one executable chunk pops off the front per step. The choice to stagger the noise levels is what makes streaming possible. A monotone staircase $\tau_1 < \cdots < \tau_N$ aligns buffer position with denoising progress and with execution order simultaneously, so the chunk most advanced in denoising is also the chunk first to execute. 
Without staggering, the buffer collapses into a batch: chunks would emerge in bulk rather than one per step, reintroducing the chunk-boundary stalls that motivated streaming.

Inside the buffer, we apply a chunk-wise causal mask: later (noisier) chunks attend to earlier (cleaner) chunks, but not the reverse. Bidirectional attention would let chunks about to execute condition on the noisier chunks behind them in execution order, violating the natural temporal direction of information flow. No inter-chunk attention would eliminate the implicit conditioning that gives the model its view of the ongoing trajectory. Chunk-wise causality lets information flow forward in time but not backward in noise.
The same buffer-and-mask design produces three runtime properties.

\textbf{Amortized denoising.} Each chunk traverses all $N$ noise levels of the buffer before execution, but the $N$ passes are spread across $N$ different forward steps, each of which also advances every other chunk in the buffer. Per-step action-decoding latency drops by up to \textbf{20$\times$} in our profile (Fig.~\ref{fig:profile}).

\textbf{Implicit asynchronous conditioning.} Because future chunks attend to near-execution chunks, the model decoding for a future robot state is already conditioned on the trajectory that state will arrive on, without an explicit future-state predictor or auxiliary action-conditioning module.

\textbf{Chunk-level memory.} Each higher-timestep chunk attends to \emph{all} earlier chunks in the buffer, not just its immediate predecessor. The buffer thus encodes a short, structured history of recently refined chunks. We hypothesize that this implicit memory accounts for \method's particularly strong gains on long-horizon tasks, and test the hypothesis in \S\ref{sec:long-horizon}.

\method's streaming design only requires slight modifications to the model's action expert architecture. We apply FiLM~\citep{film} to enhance the multi-level timestep conditioning in the action expert. For more details, please refer to Appendix~\ref{app:implementation}.

\subsection{Streaming Inference}
\label{sec:inference}

\method runs in two phases. The first fills the buffer before any prediction is executed; the second steady-state phase produces one executable chunk per forward pass.

\textbf{Cold start.} At the start of an episode, the buffer is empty and \method cannot yet emit a clean chunk. We initialize $\mathbf{B}$ with $N{-}1$ padding chunks and one Gaussian sample, then run $N{-}1$ inference steps to populate the remaining slots. During these warm-up steps the robot does not execute predicted actions; instead it executes a safe default (the initial robot state for joint-space control, or the normalized zero action for end-effector control) so that the arm remains stationary while the buffer fills. The cold-start overhead is paid once per episode and is amortized over the full rollout.

\textbf{Steady streaming.} Once the buffer is populated, \method runs as a queue (Alg.~\ref{alg:flashvla}). Each inference step (i) advances every chunk by one denoising step, (ii) pops the cleanest chunk for execution, (iii) shifts the remaining chunks forward, and (iv) appends a fresh pure-noise chunk at the tail. Each chunk traverses all $N$ noise levels of the buffer before it is executed.

\begin{table}[t]
\small\centering
\tablestyle{3pt}{1.2}
\begin{tabular}{lcccccccc}
\toprule
\textbf{Method}
& \multicolumn{5}{c}{\textbf{Success Rate (\%)}}
& \multicolumn{3}{c}{\textbf{Efficiency}} \\
\cmidrule(lr){2-6} \cmidrule(lr){7-9}
& Spatial
& Object
& Goal
& Long
& \textbf{Average}
& \textbf{Steps}
& \textbf{Time (s)}
& \textbf{Time / Step (ms)} \\
\midrule

$\pi_{0.5}$
& 98.8
& 98.2
& 98.0
& 92.4
& 96.9
& 156.0
& 8.4
& 53.8 \\

\midrule

+ VLASH (d{=}1)
& 98.8
& 99.2
& 96.7
& 94.4
& 97.2 (\textcolor{mygreen}{$\uparrow$ 0.3})
& 153.9
& 7.2
& 46.8 (\textcolor{mygreen}{1.15$\times$}) \\

+ VLASH (d{=}4)
& 92.5
& 96.9
& 93.3
& 89.6
& 93.1 (\textcolor{myred}{$\downarrow$ 3.8})
& 176.7
& 5.8
& 32.8 (\textcolor{mygreen}{1.64$\times$}) \\

+ StreamingVLA (d{=}1)
& 96.6
& 96.6
& 95.4
& 91.0
& 94.9 (\textcolor{myred}{$\downarrow$ 2.0})
& n/a
& n/a
& 31.6 (\textcolor{mygreen}{1.70$\times$}) \\

\midrule
\rowcolor{mygray}
+ \method (d{=}1)
& 98.8
& 99.6
& 97.6
& 95.4
& \textbf{97.8 (\textcolor{mygreen}{$\uparrow$ 0.9})}
& 158.0
& 3.5
& \textbf{22.1} (\textcolor{mygreen}{\textbf{2.43$\times$}}) \\

\bottomrule
\end{tabular}
\caption{Asynchronous performance on LIBERO at one-step delay ($d{=}1$) unless noted. \method reaches the largest speedup over $\pi_{0.5}$ (2.43$\times$), while improving average success rate from 96.9\% to 97.8\%. Other asynchronous baselines either provide smaller speedups or lose success rate. \textbf{Steps}: actions executed per episode (averaged over 2{,}000 episodes); \textbf{Time (s)}: per-episode completion time.}
\label{tab:libero_async}
\end{table}

\textbf{System-level optimization.} Because the per-step computation is structurally identical across steady-state iterations, the inference path is highly amenable to compilation. We compile each stage into a CUDA Graph, which is both faster to dispatch and non-blocking under asynchronous execution. We additionally pack linear layers to reduce kernel launch overhead and apply PyTorch's max-autotune compilation to remove residual kernel bubbles. These optimizations are orthogonal to the algorithmic design but are necessary to realize the full latency benefit in practice.

\subsection{Multi-Buffer Joint Fine-tuning}
\label{sec:training}

Pretrained VLAs are trained to denoise each chunk from scratch in isolation and have never seen a partially populated buffer or a chunk-wise causal attention pattern. We adapt them to the streaming paradigm during the standard downstream fine-tuning stage.

\textbf{The challenge.} For a single observation $o_t$, the cold-start procedure exposes the model to $N$ distinct buffer configurations. Configuration $j$ has $j$ real action chunks at the noisier slots and $N{-}j$ padding chunks at the cleaner slots; the final configuration $j{=}N$ is the steady-streaming state. The model must handle every such configuration. Treating each as a separate training sample multiplies the dataset by $N$, slows fine-tuning, and gives the optimizer no signal that they share a single observation.

\textbf{The packed solution.} We instead pack all $N$ buffer configurations into a single training sample (Fig.~\ref{fig:method}). Within the packed sample, all configurations attend to the same encoded observation $o_t$, but an attention mask isolates each buffer from the others. The model sees every possible buffer state at training time, while the expensive observation encoding is computed once. This packing exploits a structural fact about the streaming algorithm: the observation context is shared across buffer states; only the action portion changes.

The training objective is the standard flow-matching loss summed over all buffer configurations:
\begin{equation}
\mathcal{L}_{\text{joint}} = \sum_{j=1}^{N} \sum_{i=1}^{j} \mathbb{E}_{\tau_i, \mathbf{z}_i}\!\left[\big\|v_\theta\big(\mathbf{x}^{(i)}_{\tau_i}, \tau_i \mid o_t, \mathbf{B}^{(j)}_t\big) - (\mathbf{z}_i - \mathbf{a}^{(i)})\big\|_2^2\right].
\end{equation}
This single objective covers both cold-start and steady-streaming behavior: the model learns to denoise chunks at any noise level conditioned on any valid buffer prefix.

\section{Simulated Experiments}
\label{sec:exp}

We evaluate \method against three claims it inherits from \S\ref{sec:intro} and \S\ref{sec:method}: (i) at fixed task quality, decoding chunks jointly is substantially faster than decoding them in isolation; (ii) under asynchronous execution, chunk-wise causal attention preserves trajectory continuity without an explicit future-state module; and (iii) the chunk-level memory created by the streaming buffer translates into measurable long-horizon gains. We test (i) and (ii) jointly on the primary asynchronous-deployment benchmark (\S\ref{sec:async}), isolate (i) on a decoding-latency benchmark (\S\ref{sec:latency}), and test (iii) under synchronous execution to remove the asynchronous confound (\S\ref{sec:long-horizon}).

\begin{figure}[t]
    \centering
    \begin{subfigure}[t]{0.49\linewidth}
        \centering
        \includegraphics[width=\linewidth]{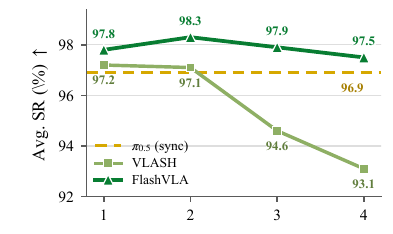}
        \caption{LIBERO success rate.}
        \label{fig:async-delay-libero-sr}
    \end{subfigure}
    \hfill
    \begin{subfigure}[t]{0.49\linewidth}
        \centering
        \includegraphics[width=\linewidth]{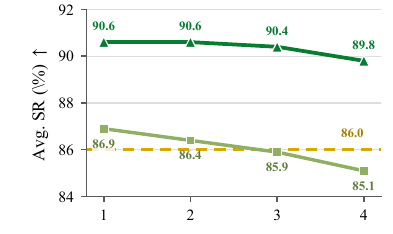}
        \caption{RoboTwin~2.0 success rate.}
        \label{fig:async-delay-robotwin-sr}
    \end{subfigure}

    \begin{subfigure}[t]{0.49\linewidth}
        \centering
        \includegraphics[width=\linewidth]{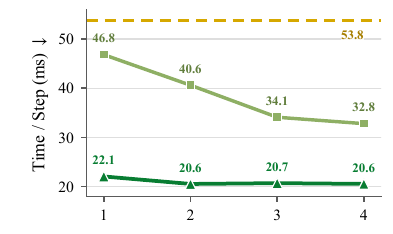}
        \caption{LIBERO time per step.}
        \label{fig:async-delay-libero-time}
    \end{subfigure}
    \hfill
    \begin{subfigure}[t]{0.49\linewidth}
        \centering
        \includegraphics[width=\linewidth]{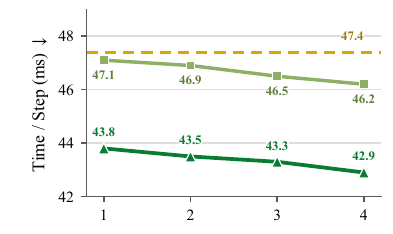}
        \caption{RoboTwin~2.0 time per step.}
        \label{fig:async-delay-robotwin-time}
    \end{subfigure}
    \caption{Success rate and time per step across asynchronous delay $d$ on LIBERO and RoboTwin~2.0. Both VLASH and \method are built on the same $\pi_{0.5}$ baseline. \method remains above the synchronous success rate across $d{=}1$--$4$ while achieving lower time per step. Success rate is averaged over four LIBERO suites and the clean and randomized evaluations of 50 RoboTwin~2.0 tasks; dashed lines denote synchronous $\pi_{0.5}$.}
    \label{fig:async-delay-sweep}
\end{figure}

\subsection{Setup}
\label{sec:exp-setup}

\textbf{Backbones and benchmarks.}
We instantiate \method primarily on $\pi_{0.5}$~\citep{pi05}, the current state-of-the-art flow-matching VLA, and evaluate its cross-architecture generalization on SmolVLA and LingBot-VLA in \S\ref{sec:cross-arch}. For simulation we use LIBERO~\citep{libero} (single-arm; four suites: Spatial, Object, Goal, Long) and RoboTwin~2.0~\citep{robotwin2.0} (bimanual; 50 tasks evaluated under both clean and randomized settings). Following LingBot-VLA~\citep{lingbot-vla}, we group the RoboTwin~2.0 tasks into short-, medium-, and long-horizon subsets for horizon-wise analysis.

\textbf{Baselines.}
For asynchronous execution we compare against VLASH~\citep{vlash} and StreamingVLA~\citep{streamingvla}, which represent the two dominant strategies for mitigating prediction-execution mismatch (future-state conditioning and action flow matching). For decoding latency we compare against Realtime-VLA~\citep{realtime-vla}, the strongest system-level acceleration baseline and an instance of the per-pass-cheaper family identified in \S\ref{sec:related}. For reaction speed we compare against FASTER~\citep{faster}.

\textbf{Configuration.}
Unless otherwise stated, \method uses chunk size 10 and buffer length 4 on LIBERO, and chunk size 20 with buffer length 4 on RoboTwin~2.0. On RoboTwin~2.0, the 50-task multitask evaluation trains one model per method on the union of clean and randomized data from all 50 tasks with a matched 16-action execution horizon; the horizon-grouped table reports the same evaluation aggregated by task horizon. Empirically, setting the total buffer span (chunk size $\times$ buffer length) close to the pretrained model's native action-chunk length yields the strongest success rates; we ablate this choice in Appendix~\ref{app:ablation}. All training is on $8\times$ H200 GPUs. Simulation runs on a single RTX 4090; the latency analysis also reports RTX 5090 numbers.

\subsection{Asynchronous Execution}
\label{sec:async}

Under asynchronous execution, the policy predicts from a stale observation and its next chunk is executed from a robot state the policy never saw. An effective VLA must therefore reduce latency without amplifying this prediction-execution mismatch. We use $d$ to denote the number of low-level action steps by which the next inference is launched early: for execution horizon $E$, inference starts after $E-d$ actions and its output is executed after the current $E$-action segment; $d{=}0$ is synchronous.

\paragraph{Overall asynchronous performance.}
At one-step asynchronous delay, \method improves both task quality and execution speed on both benchmarks. On LIBERO, it delivers the largest joint improvement over synchronous $\pi_{0.5}$, raising average success from 96.9\% to 97.8\% while reducing time per step from 53.8 to 22.1\,ms (2.43$\times$; Tab.~\ref{tab:libero_async}). VLASH provides only a 1.15$\times$ speedup at comparable success, while StreamingVLA reaches 1.70$\times$ but loses 2.0 success points. Figure~\ref{fig:async-delay-sweep} further shows that on the 50-task RoboTwin~2.0 evaluation, \method reaches 90.6\% success, compared with 86.0\% for synchronous $\pi_{0.5}$ and 86.9\% for VLASH at the same delay, while reducing time per step from 47.4 to 43.8\,ms.

\paragraph{Robustness to asynchronous delay.}
The quality and speed gains persist as the prediction-execution mismatch grows with $d$ (Fig.~\ref{fig:async-delay-sweep}). Across $d{=}1$--$4$, \method maintains 97.5--98.3\% success on LIBERO, consistently above the 96.9\% synchronous baseline, while sustaining a 2.43--2.62$\times$ speedup. In contrast, VLASH drops to 93.1\% success at $d{=}4$. On RoboTwin~2.0, \method remains 3.7--4.7 points above VLASH and above synchronous $\pi_{0.5}$ even at $d{=}4$ (89.8\% vs.\ 86.0\%), together with a 1.08--1.10$\times$ end-to-end speedup. The smaller RoboTwin~2.0 speedup reflects its simulator-dominated runtime: policy inference accounts for only 17\% of baseline step time (8.1 of 47.4\,ms), compared with 36.8\,ms for rendering and 2.5\,ms for communication. Because the execution horizon is matched throughout each sweep, varying $d$ changes observation staleness without changing the number of actions executed per segment. This robustness follows from chunk-wise causal attention, which preserves temporal consistency without external mismatch correction; removing causality while retaining the buffer collapses the asynchronous gains (Appendix~\ref{app:ablation}, Fig.~\ref{fig:causal_ablation}).

\begin{figure}[t]
\centering

\begin{minipage}[t]{0.56\linewidth}
\centering
\captionsetup{font=footnotesize}
\vspace{0pt}
\renewcommand{\arraystretch}{1.15}
\resizebox{\linewidth}{!}{
\begin{tabular}{lcccc}
\toprule
\multirow{2}{*}{\textbf{Method}}
& \multicolumn{2}{c}{\textbf{RTX 4090 (ms)}}
& \multicolumn{2}{c}{\textbf{RTX 5090 (ms)}} \\
& 2 views & 3 views
& 2 views & 3 views \\
\midrule
$\pi_{0.5}$         & 45.8 & 55.4 & 37.0 & 44.8 \\
+ Realtime-VLA      & 29.2 & 38.9 & 26.6 & 34.2 \\
\rowcolor{mygray}
+ \method           & \textbf{26.7} & \textbf{36.8} & \textbf{20.3} & \textbf{27.1} \\
\bottomrule
\end{tabular}
}
\captionof{table}{Inference latency on RTX 4090 and RTX 5090 with two- and three-view inputs (one frame per view). We average 100 samples after 10 warm-up iterations. Both $\pi_{0.5}$ and \method use the same system optimizations: CUDA Graphs and kernel fusion. \method is fastest in every configuration.}
\label{tab:latency}
\end{minipage}
\hfill
\begin{minipage}[t]{0.42\linewidth}
\centering
\captionsetup{font=footnotesize}
\vspace{0pt}
\includegraphics[width=\linewidth]{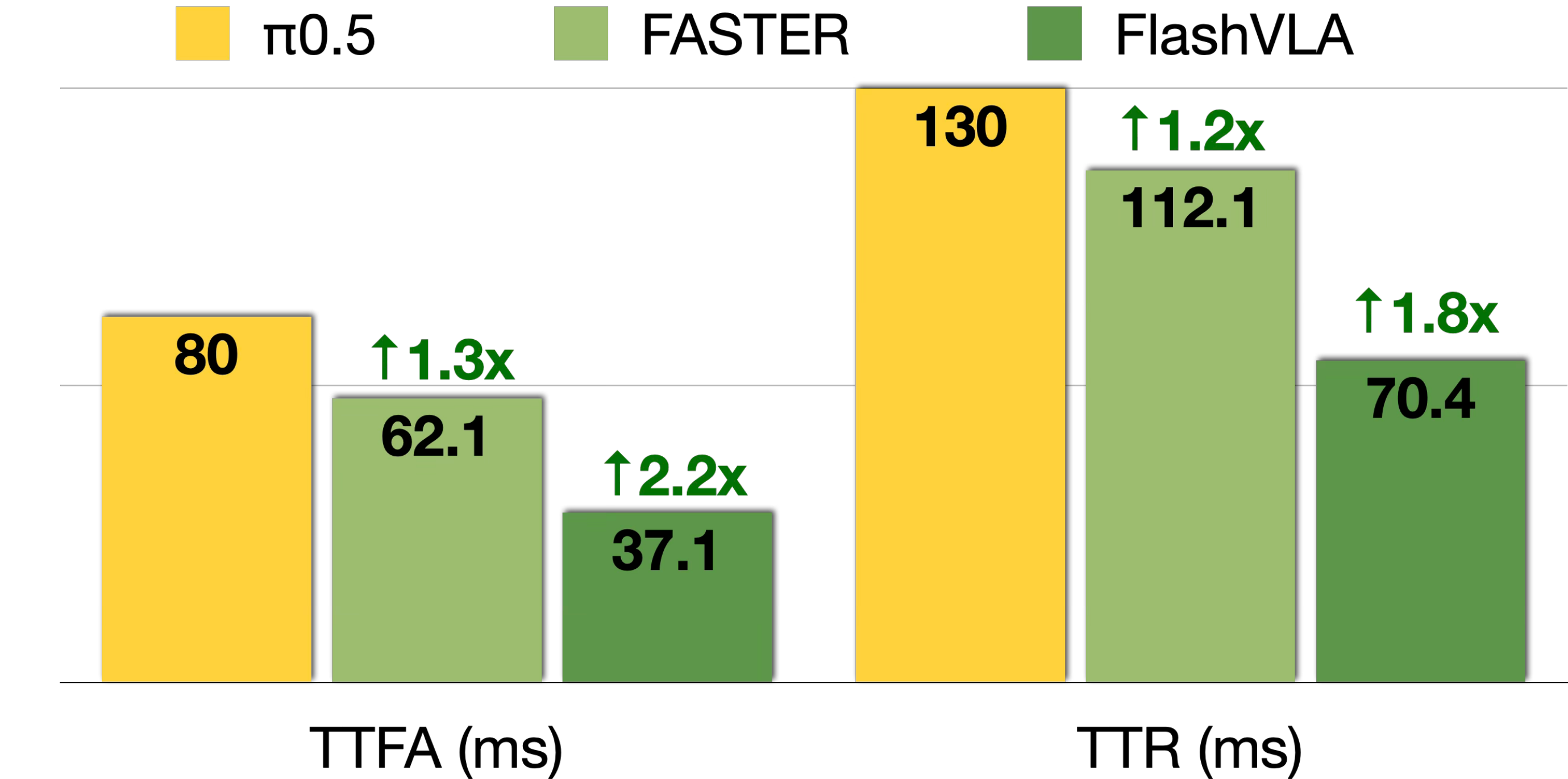}
\captionof{figure}{Reaction speed under a 30\,Hz target. \method achieves 1.7$\times$ lower time-to-first-action (TTFA) and 1.6$\times$ lower time-to-react (TTR) than FASTER.}
\label{fig:react_speed}
\end{minipage}
\end{figure}

\begin{table}[!t]
\centering
\tablestyle{3pt}{1.2}
\begin{tabular}{lcccccccc}
\toprule
\multirow{2}{*}{\textbf{Method}}
& \multicolumn{5}{c}{\textbf{LIBERO}}
& \multicolumn{3}{c}{\textbf{RoboTwin~2.0}} \\
\cmidrule(lr){2-6} \cmidrule(lr){7-9}
& Spatial & Object & Goal & Long & Avg.
& Clean & Random & Avg. \\
\midrule
$\pi_{0.5}$
& \textbf{98.8}
& 98.2
& \textbf{98.0}
& 92.4
& 96.9
& 86.1
& 85.8
& 86.0 \\
\rowcolor{mygray}
+FlashVLA
& 98.6
& \textbf{99.0}
& 97.8
& \textbf{96.2}
& \textbf{97.9 \textcolor{mygreen}{($\uparrow$ 1.0)}}
& \textbf{90.8}
& \textbf{90.2}
& \textbf{90.5 \textcolor{mygreen}{($\uparrow$ 4.5)}} \\
\bottomrule
\end{tabular}
\caption{Synchronous success rates on LIBERO and RoboTwin~2.0. On LIBERO, FlashVLA improves $\pi_{0.5}$ by 1.0 points on average, with a 3.8-point gain on LIBERO-Long. RoboTwin~2.0 reports 50-task multitask evaluation at $d{=}0$, improving the average across clean and randomized settings from 86.0\% to 90.5\%.}
\label{tab:libero_sync}
\end{table}

\subsection{Decoding Latency and Reaction Speed}
\label{sec:latency}

\begin{table}[t]
\small\centering
\tablestyle{3pt}{1.2}
\begin{tabular}{lccccccccc}
\toprule
\multirow{2}{*}{\textbf{Method}}
& \multicolumn{3}{c}{\textbf{Short Horizon}}
& \multicolumn{3}{c}{\textbf{Medium Horizon}}
& \multicolumn{3}{c}{\textbf{Long Horizon}} \\
\cmidrule(lr){2-4} \cmidrule(lr){5-7} \cmidrule(lr){8-10}
& Clean & Random & Avg.
& Clean & Random & Avg.
& Clean & Random & Avg. \\
\midrule
$\pi_{0.5}$
& \textbf{93.5} & \textbf{94.2} & \textbf{93.9}
& 81.9 & 80.4 & 81.1
& 54.2 & 51.8 & 53.0 \\
\rowcolor{mygray}
+FlashVLA
& 93.2 & 93.3 & 93.2 (\textcolor{myred}{$\downarrow$0.7})
& \textbf{86.0} & \textbf{84.8} & \textbf{85.4} (\textcolor{mygreen}{$\uparrow$4.3})
& \textbf{90.8} & \textbf{88.4} & \textbf{89.6} (\textcolor{mygreen}{$\uparrow$36.6}) \\
\bottomrule
\end{tabular}

\caption{Synchronous performance on all 50 RoboTwin~2.0 tasks, grouped by task horizon. FlashVLA preserves short-horizon performance and improves average success by 4.3 points on medium-horizon tasks and 36.6 points on long-horizon tasks.}
\label{tab:robotwin}
\end{table}

Table~\ref{tab:latency} compares per-policy-invocation inference latency with Realtime-VLA, which applies kernel-level optimization to the original decoding loop, on RTX 4090 and RTX 5090 with two- and three-camera inputs. \method is fastest in every configuration. Relative to the optimized $\pi_{0.5}$ baseline, it reduces latency from 45.8 to 26.7\,ms and from 55.4 to 36.8\,ms on the RTX 4090, and from 37.0 to 20.3\,ms and from 44.8 to 27.1\,ms on the RTX 5090, for two and three views respectively. It also outperforms Realtime-VLA in all four matched settings. The 20.3\,ms result corresponds to a sustained $\sim$50\,Hz policy update rate.

Figure~\ref{fig:react_speed} compares reaction speed against FASTER under a 30\,Hz target. We report time-to-first-action (TTFA), the per-observation policy response latency, and time-to-react (TTR), modeled by FASTER as $\Delta t_{\text{react}} \sim \mathcal{U}(\Delta t_{\text{infer}}, \Delta t_{\text{infer}} + \Delta t_{\text{exec}})$ with $\mathbb{E}[\Delta t_{\text{react}}] = \Delta t_{\text{infer}} + 0.5\,\Delta t_{\text{exec}}$. For \method we report steady-state latency after the buffer is filled; the one-time cold-start overhead is counted in the episode-level completion time and is negligible when amortized over a full rollout. \method reduces TTFA from 80.0\,ms for $\pi_{0.5}$ and 62.1\,ms for FASTER to 37.1\,ms, and expected TTR from 130.0 and 112.1\,ms to 70.4\,ms. It therefore improves both initial action availability and closed-loop reaction, achieving 1.7$\times$ lower TTFA and 1.6$\times$ lower TTR than FASTER.

\subsection{Synchronous Quality and Long-Horizon Gain}
\label{sec:long-horizon}

Under synchronous evaluation, \method improves average success from 96.9\% to 97.9\% on LIBERO (Tab.~\ref{tab:libero_sync}). On RoboTwin~2.0, it improves success from 86.1\% to 90.8\% under clean evaluation and from 85.8\% to 90.2\% under randomized evaluation (Tab.~\ref{tab:robotwin}). These results show that the streaming formulation preserves and improves synchronous policy quality across both benchmarks. Training budget and execution horizon are matched between methods, and this setting uses no prediction-execution overlap; the gain therefore isolates joint chunk decoding rather than additional optimization or more frequent replanning.

The performance gain from \method increases sharply with task horizon. Across the same 50 RoboTwin~2.0 tasks, the average difference relative to $\pi_{0.5}$ grows from $-0.7$ points on short-horizon tasks to $+4.3$ on medium-horizon tasks and $+36.6$ on long-horizon tasks; on the long-horizon subset, Clean success rises from 54.2\% to 90.8\% and Random success from 51.8\% to 88.4\%. The trend is consistent across both evaluation settings: medium-horizon gains exceed 4 points for both Clean and Random, while long-horizon gains reach 36.6 points in each. LIBERO shows the same pattern, with a 3.8-point gain on LIBERO-Long (92.4\% to 96.2\%). This is the empirical signature of the chunk-level memory described in \S\ref{sec:formulation}: under chunk-wise causal attention, each higher-noise future chunk attends to all earlier chunks in the streaming buffer, giving the action expert a short, structured history of recently refined chunks. Tasks that span more chunk transitions benefit more from this history, so the long-horizon gap is the structural advantage of the design surfacing in the metric most sensitive to it. The causal-mask ablation provides complementary evidence: removing inter-chunk attention while retaining the streaming buffer produces the largest degradation on LIBERO-Long (Fig.~\ref{fig:causal_ablation}). Although not an explicit design target, the long-horizon improvement follows directly from the chunk-wise causal formulation.

\subsection{Cross-Architecture Generalization}
\label{sec:cross-arch}
\begin{table}[t]
\centering
\tablestyle{14pt}{1.15}
\begin{tabular}{lccc}
\toprule
\textbf{Method}
& \textbf{Avg. SR (\%)}
& \textbf{Time / Step (ms)}
& \textbf{Inference Latency (ms)} \\
\midrule
& \multicolumn{3}{c}{\textit{LIBERO}} \\
\midrule
SmolVLA
& 80.1
& 41.2
& 19.7 \\
\rowcolor{mygray}
$+\,$\method{} ($d{=}0$)
& \textbf{80.1} (\textcolor{mygreen}{$\uparrow$0.0})
& \textbf{29.7} (\textcolor{mygreen}{1.39$\times$})
& \textbf{10.1} (\textcolor{mygreen}{1.95$\times$}) \\
\rowcolor{mygray}
$+\,$\method{} ($d{=}1$)
& 79.5 (\textcolor{myred}{$\downarrow$0.6})
& \textbf{28.7} (\textcolor{mygreen}{1.44$\times$})
& -- \\
\midrule
& \multicolumn{3}{c}{\textit{RoboTwin~2.0}} \\
\midrule
LingBot-VLA
& 85.2
& 56.7
& 70.6 \\
\rowcolor{mygray}
$+\,$\method{} ($d{=}0$)
& \textbf{88.6} (\textcolor{mygreen}{$\uparrow$3.4})
& \textbf{49.5} (\textcolor{mygreen}{1.15$\times$})
& \textbf{25.1} (\textcolor{mygreen}{2.81$\times$}) \\
\rowcolor{mygray}
$+\,$\method{} ($d{=}1$)
& \textbf{89.3} (\textcolor{mygreen}{$\uparrow$4.1})
& \textbf{46.6} (\textcolor{mygreen}{1.22$\times$})
& -- \\
\bottomrule
\end{tabular}
\caption{Cross-architecture generalization across SmolVLA and LingBot-VLA. With no or only minimal additional parameters, \method preserves SmolVLA success and improves LingBot-VLA by 3.4--4.1 points while reducing inference latency by 1.95$\times$--2.81$\times$. Here, $d{=}0$ denotes synchronous execution and $d{=}1$ denotes one-step asynchronous execution; delay-independent inference latency is reported only at $d{=}0$.}
\label{tab:cross_arch}
\end{table}

\method is a plug-and-play decoding method for flow-matching VLAs. Table~\ref{tab:cross_arch} shows that its policy-side acceleration transfers strongly across architectures: inference latency falls from 19.7 to 10.1\,ms on SmolVLA and from 70.6 to 25.1\,ms on LingBot-VLA, corresponding to 1.95$\times$ and 2.81$\times$ speedups. Crucially, this acceleration preserves or improves task quality. On SmolVLA, synchronous success remains at 80.1\%, while one-step asynchronous execution reaches 79.5\%, within 0.6 points of the baseline. On LingBot-VLA, success improves by 3.4 points synchronously (85.2\% to 88.6\%) and by 4.1 points asynchronously (to 89.3\%). The more moderate Time/Step gains reflect the runtime composition of the closed-loop evaluations: SmolVLA is a compact 0.5B model with only 19.7\,ms baseline inference latency, while RoboTwin~2.0 is simulator-dominated and amortizes each policy invocation over 16 executed actions. Policy inference is therefore not the dominant evaluation cost in either setting, so Time/Step does not scale one-for-one with the substantial inference-latency reduction. These results show that \method generalizes across flow-matching VLA architectures while preserving both task quality and its core policy-side efficiency benefit. Full settings are in Appendix~\ref{app:settings}. The adaptation is lightweight: it modifies the action expert's timestep conditioning and attention mask, reuses native FiLM layers when available, and adds lightweight time-MLP and FiLM layers only for SmolVLA before multi-buffer fine-tuning (Appendix~\ref{app:implementation}).

\section{Real-World Deployment}
\label{sec:realworld}

We deploy \method on a 7-DoF Franka arm with a single RTX A4000 GPU and ask whether the simulated latency and asynchronous-continuity benefits transfer to a real robot at 30\,Hz control.

\paragraph{Setup.}
We evaluate three tasks chosen to span the horizon range observed in simulation: \emph{pick and place} (short: pick a fruit, place it in a basket), \emph{whiteboard wiping} (medium: pick up a brush, wipe all marks), and \emph{table cleaning} (long: clean three objects one by one into a basket). For each task we collect 50 teleoperated demonstrations with Gello~\citep{gello}, yielding 22K, 30K, and 68K frames respectively. We compare four configurations: $\pi_{0.5}$ synchronous, $\pi_{0.5}$ naive asynchronous, $\pi_{0.5}$ asynchronous with RTC~\citep{rtc}, and \method asynchronous. All four configurations use the same CUDA Graph and kernel-fusion system optimizations, and all asynchronous methods use a two-step delay. Each task is run for 15 trials; performance is scored on a three-point rubric (2 for full success, 1 for partial completion, 0 for failure), and we additionally report the average completion time over successful trials. Full training and evaluation details are in Appendix~\ref{app:realworld_settings}.

\begin{figure}[H]
    \centering
    \begin{subfigure}[t]{0.32\linewidth}
        \centering
        \includegraphics[width=\linewidth]{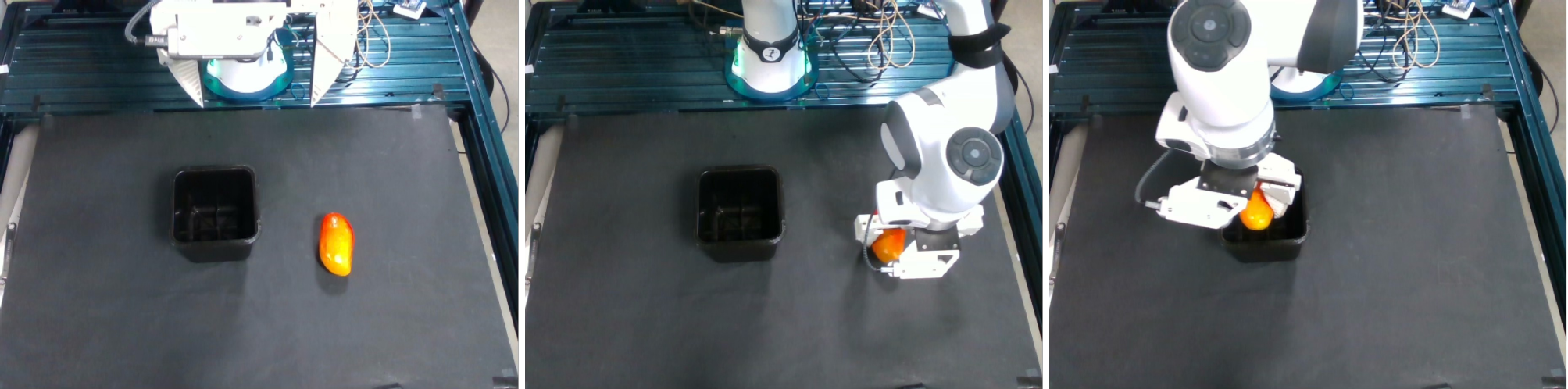}
        \caption{Pick and place (short).}
    \end{subfigure}
    \hfill
    \begin{subfigure}[t]{0.32\linewidth}
        \centering
        \includegraphics[width=\linewidth]{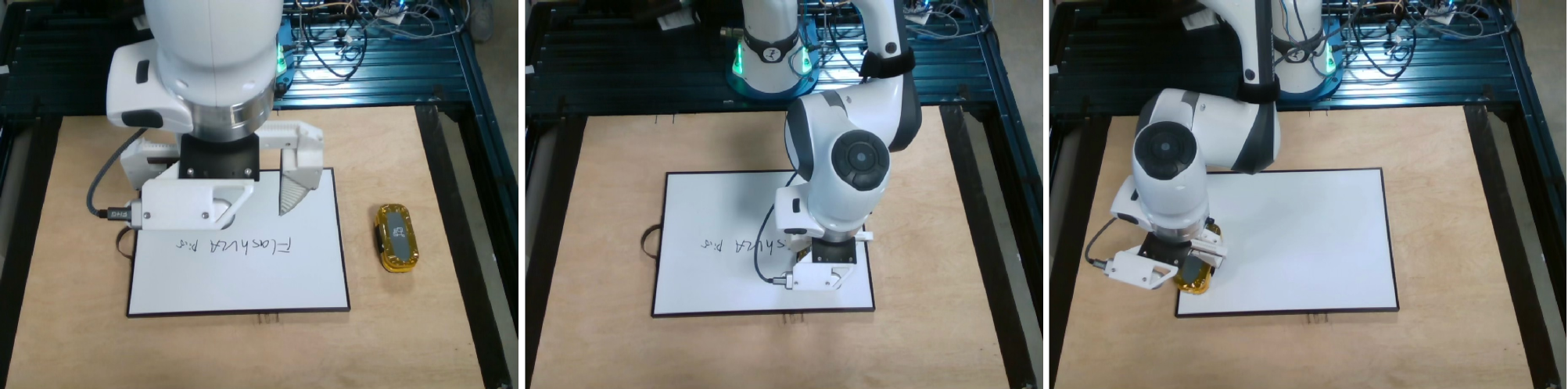}
        \caption{Whiteboard wiping (medium).}
    \end{subfigure}
    \hfill
    \begin{subfigure}[t]{0.32\linewidth}
        \centering
        \includegraphics[width=\linewidth]{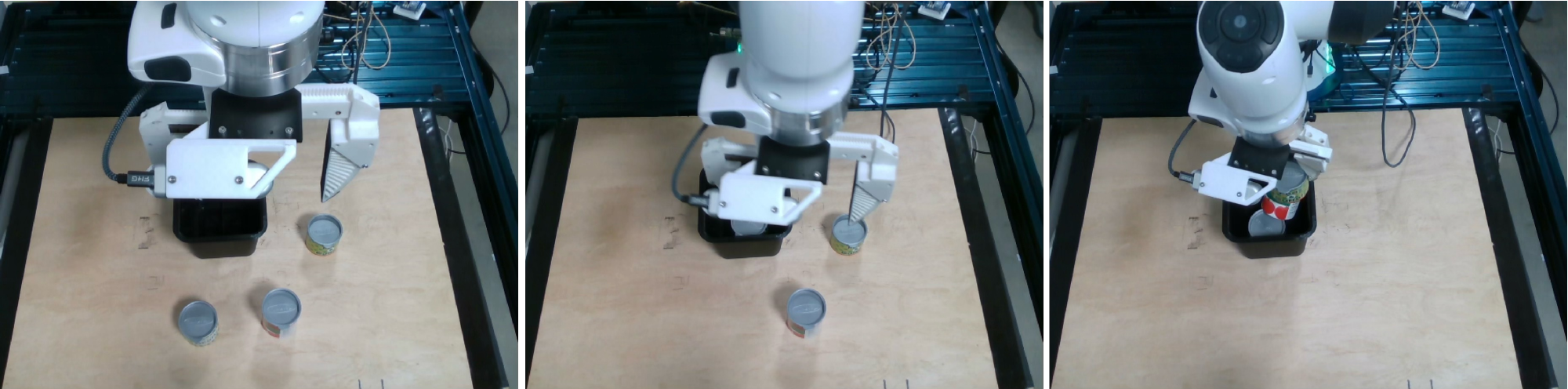}
        \caption{Table cleaning (long).}
    \end{subfigure}

    \vspace{0.4em}

    \begin{subfigure}[t]{0.48\linewidth}
        \centering
        \includegraphics[width=\linewidth]{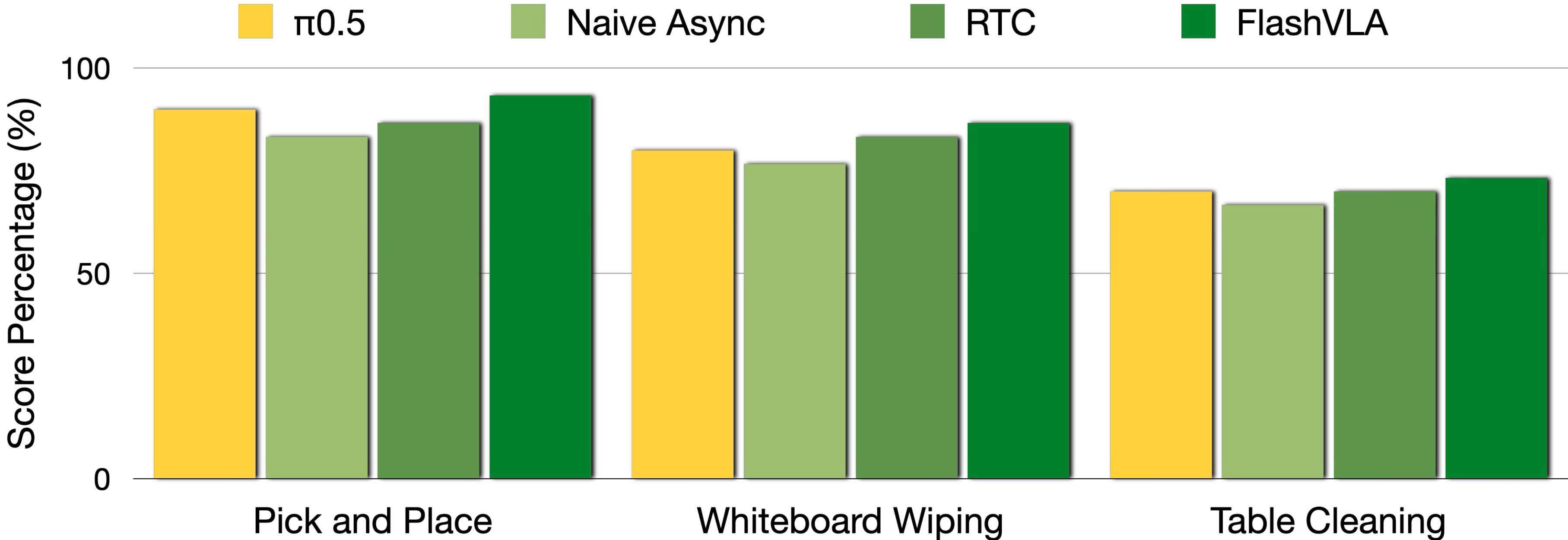}
        \caption{Task score.}
    \end{subfigure}
    \hfill
    \begin{subfigure}[t]{0.48\linewidth}
        \centering
        \includegraphics[width=\linewidth]{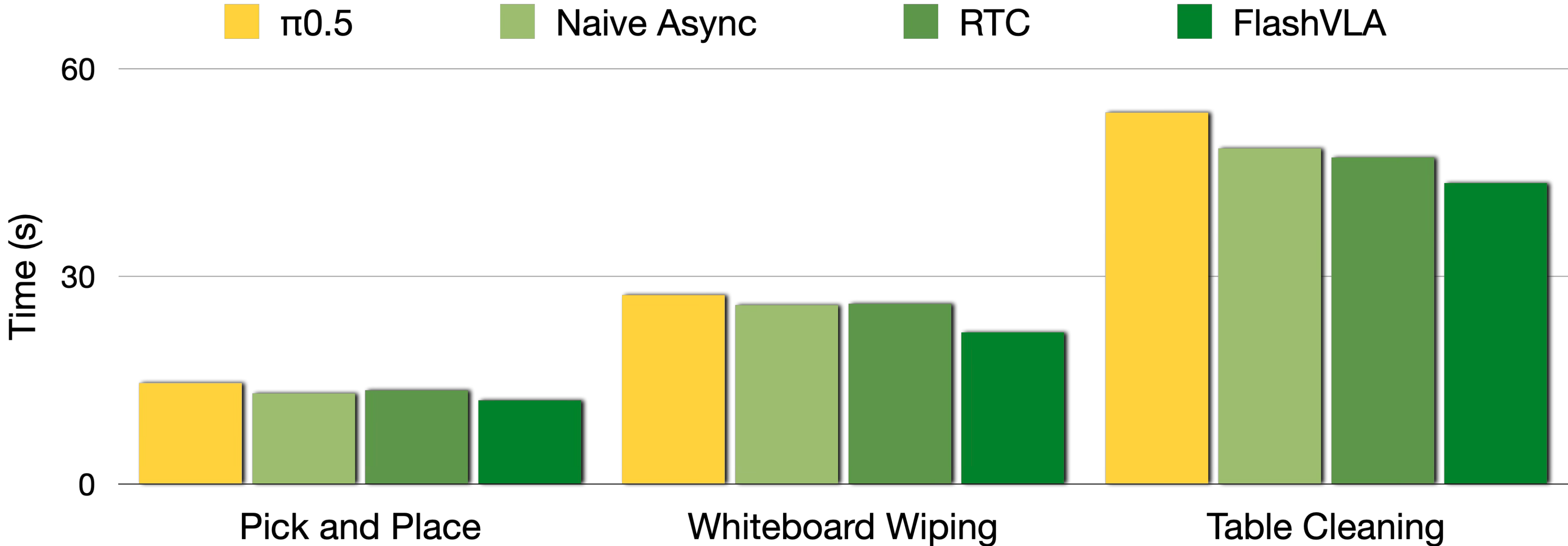}
        \caption{Completion time.}
    \end{subfigure}
    \caption{Real-world tasks and results. Top: the short-, medium-, and long-horizon tasks used for evaluation. Bottom: task score and average completion time over successful trials. \method outperforms all three $\pi_{0.5}$ configurations on every task and averages a 1.3$\times$ completion-time speedup over synchronous execution.}
    \label{fig:realworld_results}
\end{figure}

\paragraph{Results.}
Figure~\ref{fig:realworld_results} reports task score and completion time across the three tasks. \method outperforms all three $\pi_{0.5}$ configurations on every task in both metrics, improving the cross-task average score from 80.0\% for synchronous $\pi_{0.5}$ and RTC (75.6\% for naive asynchronous execution) to 84.4\%. The completion-time speedup over $\pi_{0.5}$ synchronous averages 1.3$\times$ and over RTC averages 1.2$\times$. Since completion time is measured only over successful trials, this speedup does not count early termination on failed rollouts; the joint gain in score and speed instead shows that streaming reduces chunk-boundary stalls while preserving asynchronous continuity. The speedup also grows with task horizon, mirroring the long-horizon pattern observed in simulation (\S\ref{sec:long-horizon}): longer tasks contain more chunk transitions over which steady streaming can accumulate its benefit. On the RTX A4000, \method can achieve an inference latency of 67.3\,ms, approximately two 30\,Hz control periods; the two-step asynchronous delay overlaps this latency with action execution, enabling 30\,Hz control.

\section{Conclusion}

We present \method, a streaming action decoder for flow-matching VLAs. A single design choice (a streaming buffer of action chunks at staggered noise levels with chunk-wise causal attention) collapses the latency cost and the asynchronous-mismatch cost that motivated two separate lines of prior work. Across LIBERO, RoboTwin~2.0, and real-world tasks, \method delivers up to 20$\times$ lower per-step action-decoding latency, recovers asynchronous continuity without an explicit future-state predictor, and sustains $\geq$30\,Hz smooth control on a single GPU. The chunk-wise causal mask is the active ingredient: removing it while keeping the streaming buffer collapses the asynchronous gains. The design transfers streaming chunk-wise diffusion from long video into robot action decoding, where lower-noise chunks in the buffer are not just about to be shown but about to be executed; we expect this transfer to generalize to other streaming-friendly heads in the VLA stack.

\paragraph{Limitations and future work.}
\method adapts pretrained VLAs through multi-buffer joint fine-tuning, which makes it a drop-in modification but inherits the pretrained model's independent-chunk training objective; pretraining a VLA from scratch under the chunk-wise causal formulation is a natural next step and likely to yield further gains. The streaming buffer also requires a one-time cold-start of $N{-}1$ warm-up steps per episode; the cost amortizes quickly over multi-second rollouts but is more noticeable on very short tasks.

\section*{Acknowledgment}
We gratefully acknowledge Professor Xiaolong Wang's lab for providing a Franka robot arm.

\clearpage

\bibliography{reference}  

\clearpage
\appendix
\section{Appendix}

\subsection{Simulated Experimental Settings}
\label{app:settings}
\paragraph{LIBERO.}
We fine-tune the model on the merged training dataset from HuggingFaceVLA and use mean-std statistics for action normalization. We set the learning rate to $10^{-4}$ and use a cosine scheduler with 1,000 warm-up steps, training for 50K steps in total. The per-GPU batch size is 32, and training is performed on 8 H200 GPUs. For $\pi_{0.5}$, the action chunk size is 50. For FlashVLA, we use a chunk size of 10 and a buffer length of 4.
Evaluation is conducted on an RTX 4090 using LeRobot's LIBERO evaluation script. The model replans every 5 action steps. Time/Step is the closed-loop average episode wall time divided by the average number of executed low-level action steps, including rendering time. The LIBERO results in Tab.~\ref{tab:libero_async} and Tab.~\ref{tab:libero_sync} are averaged across five runs.

\paragraph{RoboTwin~2.0.}
The training data is constructed using RoboTwin~2.0's official scripts, converted into LeRobot's format, and normalized with 99th percentile statistics. For the 50-task multitask results in Fig.~\ref{fig:async-delay-sweep} and the horizon-grouped results in Tab.~\ref{tab:robotwin}, we train one model per method on the union of clean and randomized training data from all 50 tasks. Both $\pi_{0.5}$ and \method use a learning rate of $5 \times 10^{-5}$ with cosine decay, a global batch size of 256 (8 per GPU on 8 H200 GPUs with four gradient-accumulation steps), and 100K optimization steps, corresponding to approximately 4.5 epochs. For $\pi_{0.5}$, the action chunk size is 50. For FlashVLA, we use a chunk size of 20 and a buffer length of 4.
Evaluation uses RoboTwin~2.0's official evaluation scripts and RTX 4090 configuration. Both methods use a 16-action execution horizon; $\pi_{0.5}$ therefore replans after the first 16 actions of its 50-action output. TPS follows the same closed-loop definition as LIBERO and additionally includes client-server communication. Table~\ref{tab:robotwin} aggregates this evaluation according to the task-horizon categorization of LingBot-VLA~\citep{lingbot-vla}.

\paragraph{Cross-architecture evaluation.}
For SmolVLA, we fine-tune from \texttt{lerobot/smolvla\_base} with chunk size 10 and buffer length 5 for 20K steps, using AdamW with a learning rate of $10^{-4}$, cosine decay, 1,000 warm-up steps, a per-GPU batch size of 64, and no gradient accumulation. SmolVLA is evaluated on all four LIBERO suites with a 5-action execution horizon. For LingBot-VLA on RoboTwin~2.0, we use the 4B backbone with chunk size 20 and buffer length 4 and fine-tune on the unified clean and randomized data from all 50 tasks for 100K steps. We use AdamW with a learning rate of $5 \times 10^{-5}$, cosine decay, 5,000 warm-up steps, and a global batch size of 256 (8 per GPU on 8 GPUs with four gradient-accumulation steps). LingBot-VLA is evaluated with RoboTwin~2.0's official scripts under a 16-action execution horizon, and success rate is averaged over the clean and randomized settings. All cross-architecture evaluations run on a single RTX 4090. Inference latency is averaged over 100 samples after 10 warm-up iterations; each baseline and its \method variant use the same CUDA Graph and kernel-fusion optimizations.

\subsection{Real-World Experiment Settings}
\label{app:realworld_settings}
We fine-tune the model on human-collected data for each real-world task and use mean-std
for normalization. We set the learning rate to $5 \times 10^{-5}$ and use a cosine scheduler with 1,000 warm-up steps. For pick-and-place, we train for 10K steps. For whiteboard wiping and table cleaning, we train for 30K steps. The per-GPU batch size is 16, and training is performed on 8 H200 GPUs. For $\pi_{0.5}$, the action chunk size is 50. For FlashVLA, we use a chunk size of 20 and a buffer length of 4.

We deploy the fine-tuned model on an RTX A4000 at a 30~Hz control frequency. All methods use the same CUDA Graph and kernel-fusion system optimizations. We set the execution horizon to 16 and the asynchronous delay to two steps for all asynchronous methods. Each task score is averaged over 15 rollouts; completion time is averaged over successful rollouts only.

\begin{table}[H]
\centering
\renewcommand{\arraystretch}{1.2}
\resizebox{\linewidth}{!}{
\begin{tabularx}{\linewidth}{l|XXX}
\toprule
\textbf{Task} & Pick and Place & Whiteboard Wiping & Table Cleaning \\
\midrule
\textbf{Instruction}
& Pick up the fruit and place it into the basket
& Wipe all writings off the whiteboard
& Clean the table by putting all objects into the basket \\
\bottomrule
\end{tabularx}
}
\caption{Real-world tasks and the natural-language instructions provided to the model.}
\label{tab:realworld_tasks}
\vspace{-1em}
\end{table}

\subsection{Ablations and Additional Results}
\label{app:ablation}

\paragraph{Chunk-wise causal attention is the active ingredient.}
\begin{figure}[t]
    \centering
    \includegraphics[width=0.9\linewidth]{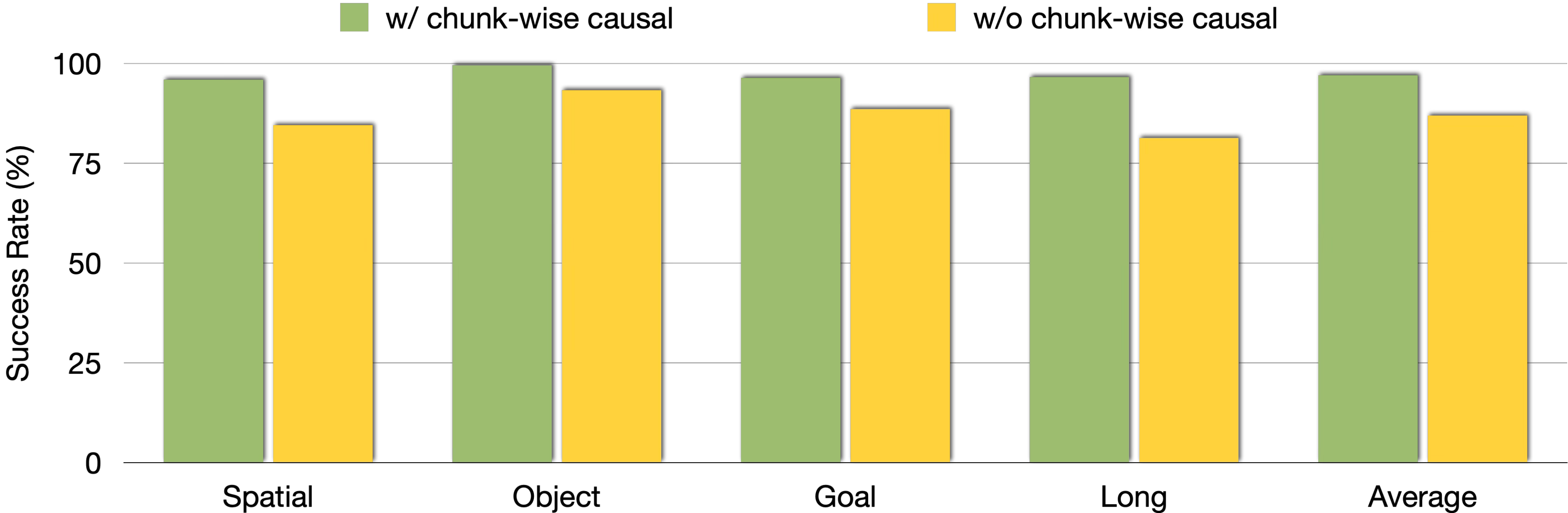}
    \caption{Chunk-wise causal attention ablation on LIBERO. Removing the causal mask while keeping the same streaming buffer drops asynchronous success rate by roughly 10 points on average, isolating the chunk-wise causal mask as the active ingredient in \method's asynchronous continuity.}
    \label{fig:causal_ablation}
\end{figure}

The most direct test of the thesis in \S\ref{sec:method} is whether \method's asynchronous continuity depends on the chunk-wise causal mask or merely on the streaming buffer. We construct a no-causal variant that retains the streaming buffer, chunk size, buffer length, training pipeline, and inference loop, but removes the chunk-wise causal mask so chunks in the buffer no longer attend to each other in any direction. As shown in Fig.~\ref{fig:causal_ablation}, the asynchronous success rate drops by roughly 10 points on average. The streaming buffer alone is therefore not sufficient; the causal attention from cleaner to noisier chunks is what makes asynchronous execution lossless under lookahead.

\paragraph{Chunk size.}
\begin{figure}[t]
    \begin{subfigure}[t]{0.48\linewidth}
    \centering
    \includegraphics[width=\linewidth]{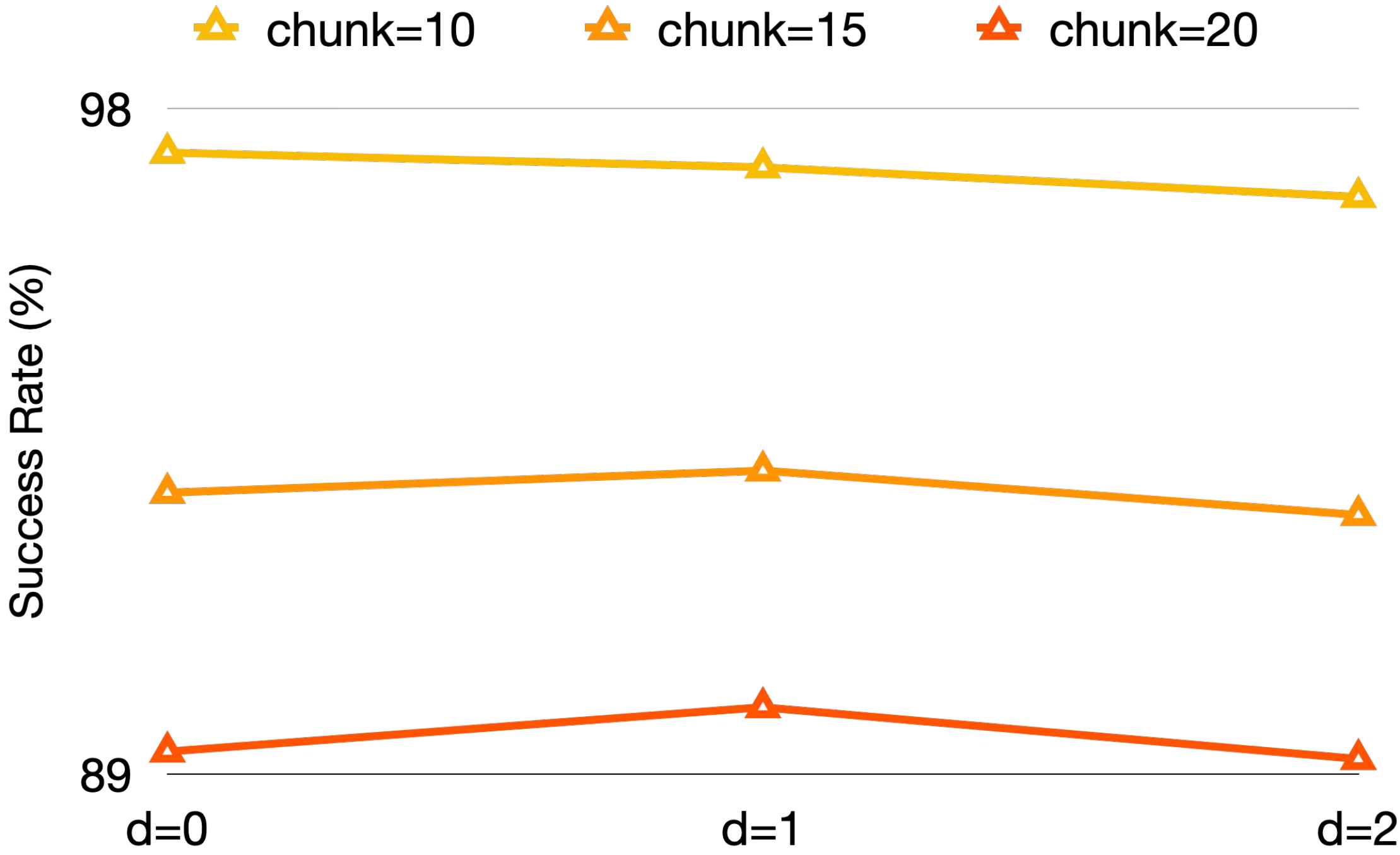}
    \caption{Success rate decreases with chunk size $C$; one-step async beats synchronous at large $C$ ($C{=}15$ and $C{=}20$).}
    \end{subfigure}
    \hfill
    \begin{subfigure}[t]{0.48\linewidth}
        \centering
        \includegraphics[width=\linewidth]{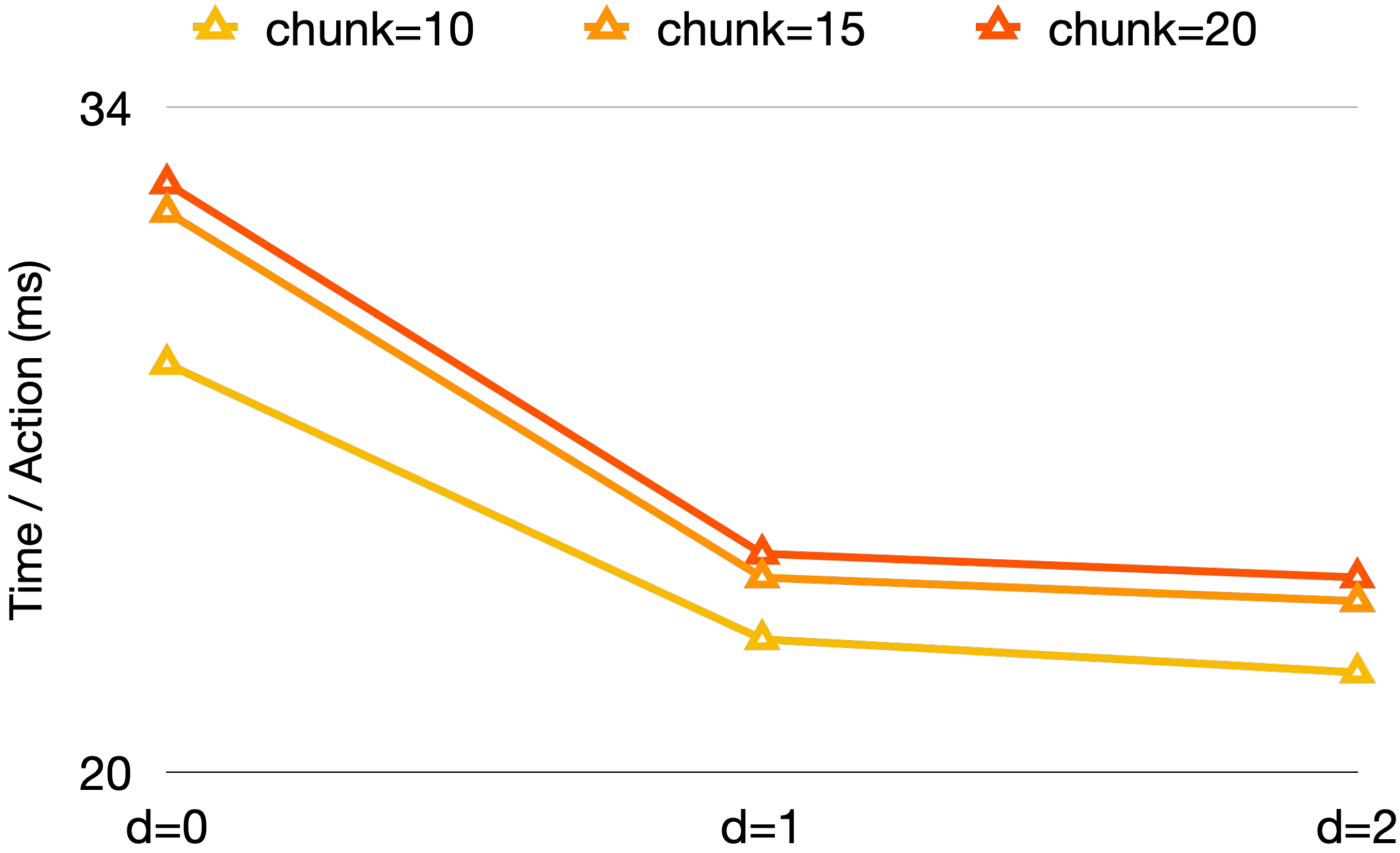}
        \caption{Per-step latency. The apparent improvement at larger $C$ is an artifact of lower success rates triggering more reset steps per episode; the genuine one-step async speedup over $\pi_{0.5}$ averages 1.8$\times$.}
    \end{subfigure}
    \caption{Chunk size $C$ ablation on LIBERO with buffer length $N{=}5$ fixed. (a) Success rate. (b) Per-step latency.}
    \label{fig:chunk_size}
\end{figure}

We sweep chunk size $C$ on LIBERO with buffer length $N{=}5$ fixed. Figure~\ref{fig:chunk_size} shows $C{=}10$ is the sweet spot under all of synchronous, one-step async, and two-step async execution. Larger chunk sizes degrade success rate; the apparent improvement in per-action time at larger $C$ is an artifact of lower success rates triggering more reset steps per episode. The pattern is consistent with the design guideline that the total buffer span $N \times C$ should approximately match the pretrained model's native action-chunk length (50 for $\pi_{0.5}$). With $N{=}5$, this prefers $C{=}10$; pushing $C$ to 15 or 20 forces the model to predict up to 100 future actions per buffer, which the short-horizon LIBERO setting does not train for. Even at $C{=}15$ and $C{=}20$, one-step async outperforms synchronous, indicating the streaming advantage persists well outside the sweet spot.

\paragraph{Buffer length.}
\begin{figure}[t]
    \begin{subfigure}[t]{0.48\linewidth}
    \centering
    \includegraphics[width=\linewidth]{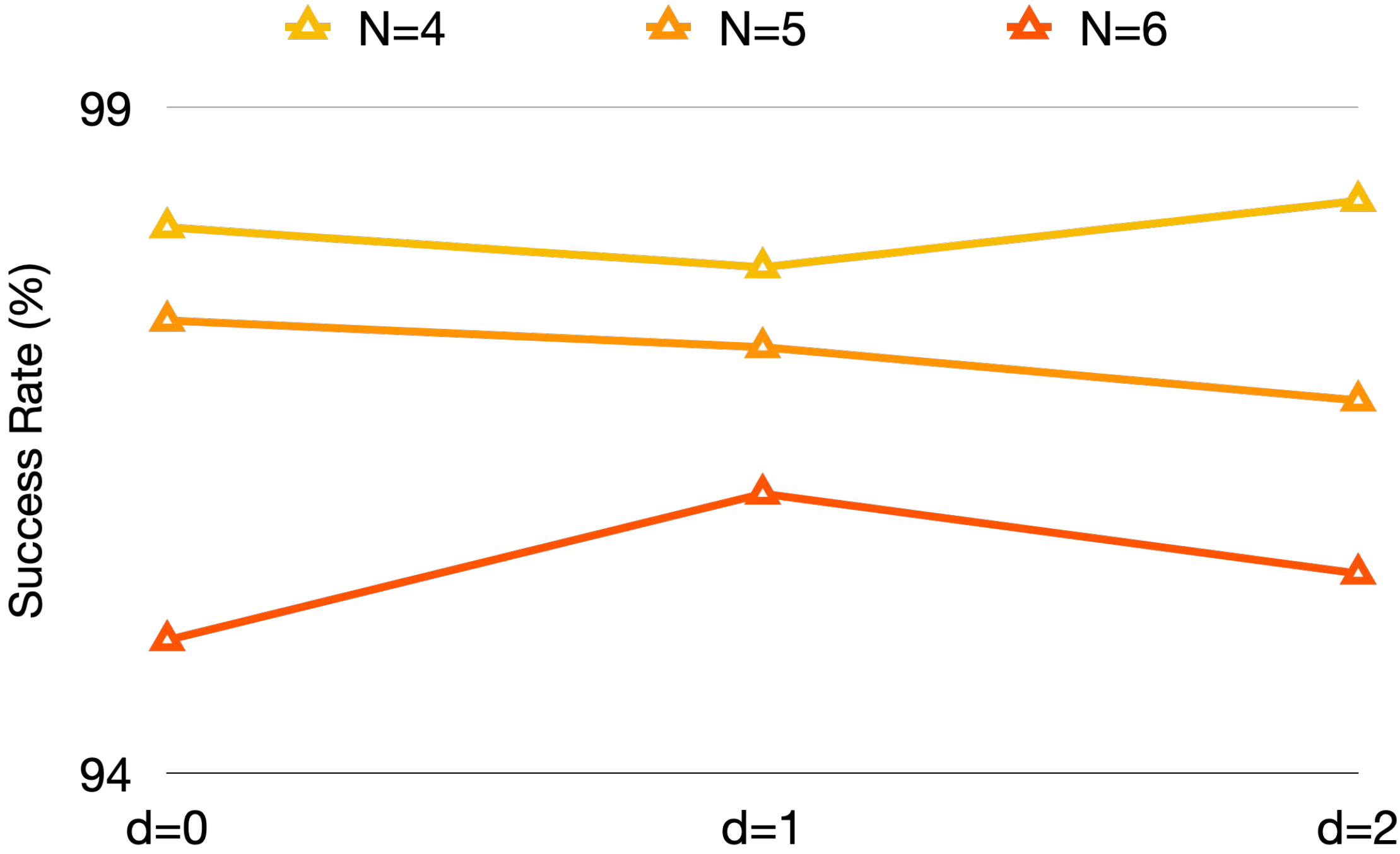}
    \caption{Success rate is robust to buffer length $N$: one-step async matches or exceeds the $\pi_{0.5}$ baseline for $N \in \{4, 5, 6\}$.}
    \end{subfigure}
    \hfill
    \begin{subfigure}[t]{0.48\linewidth}
        \centering
        \includegraphics[width=\linewidth]{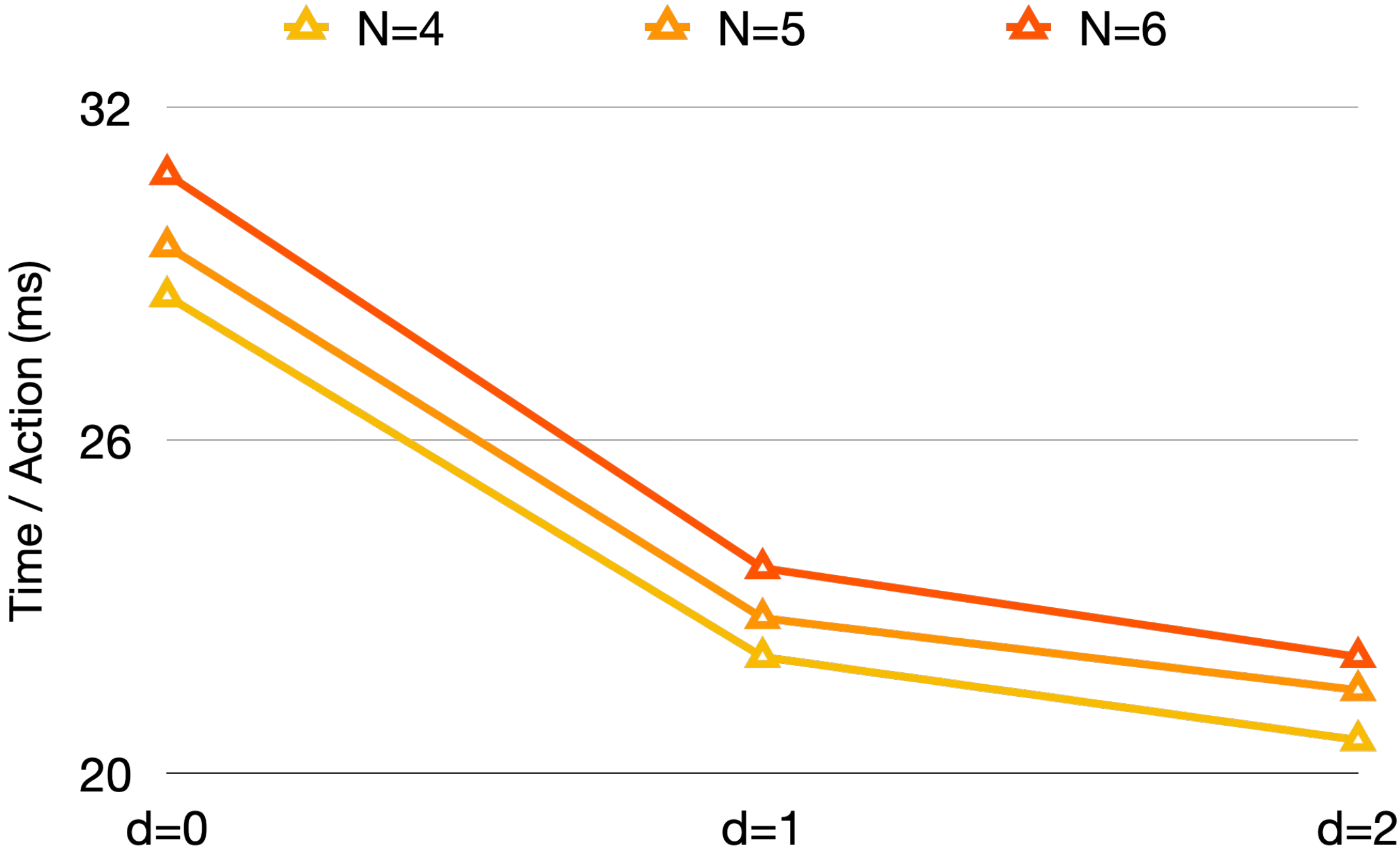}
        \caption{Per-step latency is also stable across $N$; one-step async averages 1.8$\times$ speedup over $\pi_{0.5}$ in every configuration.}
    \end{subfigure}
    \caption{Buffer length $N$ ablation on LIBERO with chunk size $C{=}10$ fixed. (a) Success rate. (b) Per-step latency.}
    \label{fig:buffer_length}
\end{figure}

Figure~\ref{fig:buffer_length} sweeps the buffer length $N \in \{4, 5, 6\}$ with $C{=}10$ fixed. \method is largely insensitive to $N$, especially under asynchronous execution: all three settings match or exceed the $\pi_{0.5}$ baseline at one-step delay, and the average speedup is 1.8$\times$ in every case. Increasing $N$ deepens the staggered noise schedule, while the observed per-step latency remains nearly unchanged over this range. Practically, $N{=}4$ is a stable default on both LIBERO and RoboTwin~2.0.

\subsection{Implementation Details}
\label{app:implementation}
\paragraph{Model architecture.}
Through experiments, we find that strengthening timestep conditioning in the action expert leads to substantially better performance. Specifically, we use FiLM~\citep{film} to generate scale, shift, and gate parameters from the multi-level timestep embeddings. This is natural for models such as $\pi_{0.5}$, where FiLM-style conditioning is already implemented. For SmolVLA, which injects timestep information through a linear projection of the concatenated action and time embeddings, we additionally introduce time-MLP and FiLM layers to enhance multi-level timestep conditioning. Also, if the model explicitly uses a state embedding, we add the time and state embeddings together as the FiLM condition. \method is fine-tuned from pretrained VLA weights rather than trained from scratch. Since the streaming formulation changes the action expert's timestep-conditioning distribution from a single denoising timestep to multi-level chunk-wise timesteps, fine-tuning can occasionally trigger gradient explosion in the pretrained action expert. Our diagnosis localizes this instability to normalization layers in the action expert: during training, the outputs of some norm layers can become numerically unstable. On datasets where this occurs, we re-initialize the action expert's normalization layers and train them from scratch while loading the remaining pretrained weights; we use this strategy for RoboTwin~2.0 50-task multitask training.

\paragraph{Multi-buffer joint fine-tuning.}
FlashVLA introduces multi-buffer joint fine-tuning to adapt pretrained VLA models to the streaming action decoding paradigm. For an observation at timestamp $t$, we retrieve the future action chunks
$a_{t:t+C}$, $a_{t+C:t+2C}$, $\dots$, $a_{t+(N-1)C:t+NC}$,
where $C$ is the chunk size and $N$ is the buffer length, i.e., the number of different flow-matching timesteps. For the $j$-th buffer state, we construct the action buffer as
\[
    \left[
    \underbrace{\text{pad}, \dots, \text{pad}}_{N-j\ \text{padding chunks}},
    \underbrace{
    a_{t:t+C},\,
    a_{t+C:t+2C},\,
    \dots,\,
    a_{t+(j-1)C:t+jC}
    }_{j\ \text{real action chunks}}
    \right],
    \quad j=1,\dots,N.
\]
We then pack all $N$ buffer states into a single training sample, where they share the same observation $o_t$ but are separated by attention masks to prevent cross-buffer interaction. We find that this multi-buffer fine-tuning strategy is crucial for enabling pretrained VLA models to efficiently adapt to FlashVLA's streaming action decoding paradigm.

\paragraph{Streaming inference.}
We provide \method's streaming inference algorithm in Algorithm~\ref{alg:flashvla}. The streaming inference consists of two stages: cold-start and steady-streaming.

\begin{algorithm}[t]
\caption{FlashVLA Streaming Inference}
\label{alg:flashvla}
\begin{algorithmic}[1]
\Require Observation stream $\{o_t\}$, buffer length $N$, cold-start action $a_{\mathrm{cs}}$
\State Initialize buffer $\mathbf{B} \leftarrow [P,\dots,P,\mathbf{z}]$, where $\mathbf{z}\sim\mathcal{N}(0,I)$

\Comment{Cold-start: fill the buffer without executing predictions}
\For{$i=1$ to $N-1$}
    \State $\mathbf{B} \leftarrow \mathrm{FlashVLA}(\mathbf{B}, o_t)$
    \State Execute $a_{\mathrm{cs}}$
    \State $\mathbf{B} \leftarrow \mathrm{PushBack}(\mathbf{B}, \mathbf{z})$, where $\mathbf{z}\sim\mathcal{N}(0,I)$
\EndFor

\Comment{Steady streaming: output one chunk per inference}
\While{episode not terminated}
    \State $\mathbf{B} \leftarrow \mathrm{FlashVLA}(\mathbf{B}, o_t)$
    \State $\hat{\mathbf{a}} \leftarrow \mathrm{PopFront}(\mathbf{B})$
    \State Execute $\hat{\mathbf{a}}$
    \State $\mathbf{B} \leftarrow \mathrm{PushBack}(\mathbf{B}, \mathbf{z})$, where $\mathbf{z}\sim\mathcal{N}(0,I)$
\EndWhile
\end{algorithmic}
\end{algorithm}


\end{document}